\documentclass[lettersize,journal]{IEEEtran}
\usepackage{amsmath,amsfonts}
\usepackage{algorithmic}
\usepackage{array}
\usepackage[caption=false,font=normalsize,labelfont=sf,textfont=sf]{subfig}
\usepackage{textcomp}
\usepackage{stfloats}
\usepackage{url}
\usepackage{verbatim}
\usepackage{graphicx}
\usepackage{cite}
\def\BibTeX{{\rm B\kern-.05em{\sc i\kern-.025em b}\kern-.08em
		T\kern-.1667em\lower.7ex\hbox{E}\kern-.125emX}}
\usepackage{balance}
\usepackage{color}
\usepackage{multirow}
\usepackage{booktabs}

\newcommand{\etal}{\textit{et al}.\:}

\title{A Joint Visual Compression and Perception Framework for Neuralmorphic Spiking Camera}

\author{Kexiang~Feng,
	Chuanmin~Jia,
	Siwei~Ma,~\IEEEmembership{Fellow,~IEEE,}
	and~Wen~Gao,~\IEEEmembership{Fellow,~IEEE}
	\thanks{K. Feng is with Institute of Computing Technology, Chinese Academy of Sciences,
		Beijing 100190, China, with the University of Chinese Academy of Sciences, Beijing 100049, China,
		and also with the National Engineering Research Center of Visual Technology, Peking University, Beijing 100871, China
		(e-mail: fengkexiang21@mails.ucas.ac.cn).}
	\thanks{C. Jia is with the Wangxuan Institute of Computer Technology, Peking University, Beijing 100080, China (email: cmjia@pku.edu.cn).}
	\thanks{S. Ma and W. Gao are with the National Engineering Research Center of Visual Technology, Peking University, Beijing 100871, China (email: \{swma,wgao\}@pku.edu.cn).}
	\thanks{Correspondence to: Chuanmin Jia.}
	}

\begin{document}	
	\IEEEtitleabstractindextext{%
		\begin{abstract}
			The advent of neuralmorphic spike cameras has garnered significant attention for their ability to capture continuous motion with unparalleled temporal resolution.
			However, this imaging attribute necessitates considerable resources for binary spike data storage and transmission.
			In light of compression and spike-driven intelligent applications, we present the notion of Spike Coding for Intelligence (\textbf{SCI}), wherein spike sequences are compressed and optimized for both bit-rate and task performance.
			Drawing inspiration from the mammalian vision system, we propose a dual-pathway architecture for separate processing of spatial semantics and motion information, which is then merged to produce features for compression.
			A refinement scheme is also introduced to ensure consistency between decoded features and motion vectors.
			We further propose a temporal regression approach that integrates various motion dynamics, capitalizing on the advancements in warping and deformation simultaneously.
			Comprehensive experiments demonstrate our scheme achieves state-of-the-art (SOTA) performance for spike compression and analysis.
			We achieve an average \textbf{17.25\%} BD-rate reduction compared to SOTA codecs and a \textbf{4.3\%} accuracy improvement over SpiReco for spike-based classification, with 88.26\% complexity reduction and 42.41\% inference time saving on the encoding side.
		\end{abstract}
		
		}

	\maketitle
	\IEEEdisplaynontitleabstractindextext
	\IEEEpeerreviewmaketitle	
	
	\section{Introduction}
	\IEEEPARstart{T}{he} surge in autonomous driving, UAV technologies, and intelligent city strategies has intensified the demand for scene capture and imaging of ultra-high motion dynamics.
	These applications operate in high-speed environments, requiring a high temporal resolution to enable high-frame rate imaging and real-time identification, posing a significant challenge for traditional cameras.
	The target object must remain stationary during the rolling shutter scanning period to avoid motion blur and structural distortion~\cite{meilland2013unified}.
	Furthermore, due to the synchronous imaging principle of the device, there may be instances where certain content or information occurring between two successive shots may not be accurately recorded~\cite{schoberl2012photometric}.
	These limitations fail to meet the specific requirements of the aforementioned applications~\cite{litzenberger2006embedded}\cite{moeslund2006survey}.
	\begin{figure}
		\centering
		\includegraphics[width=\linewidth]{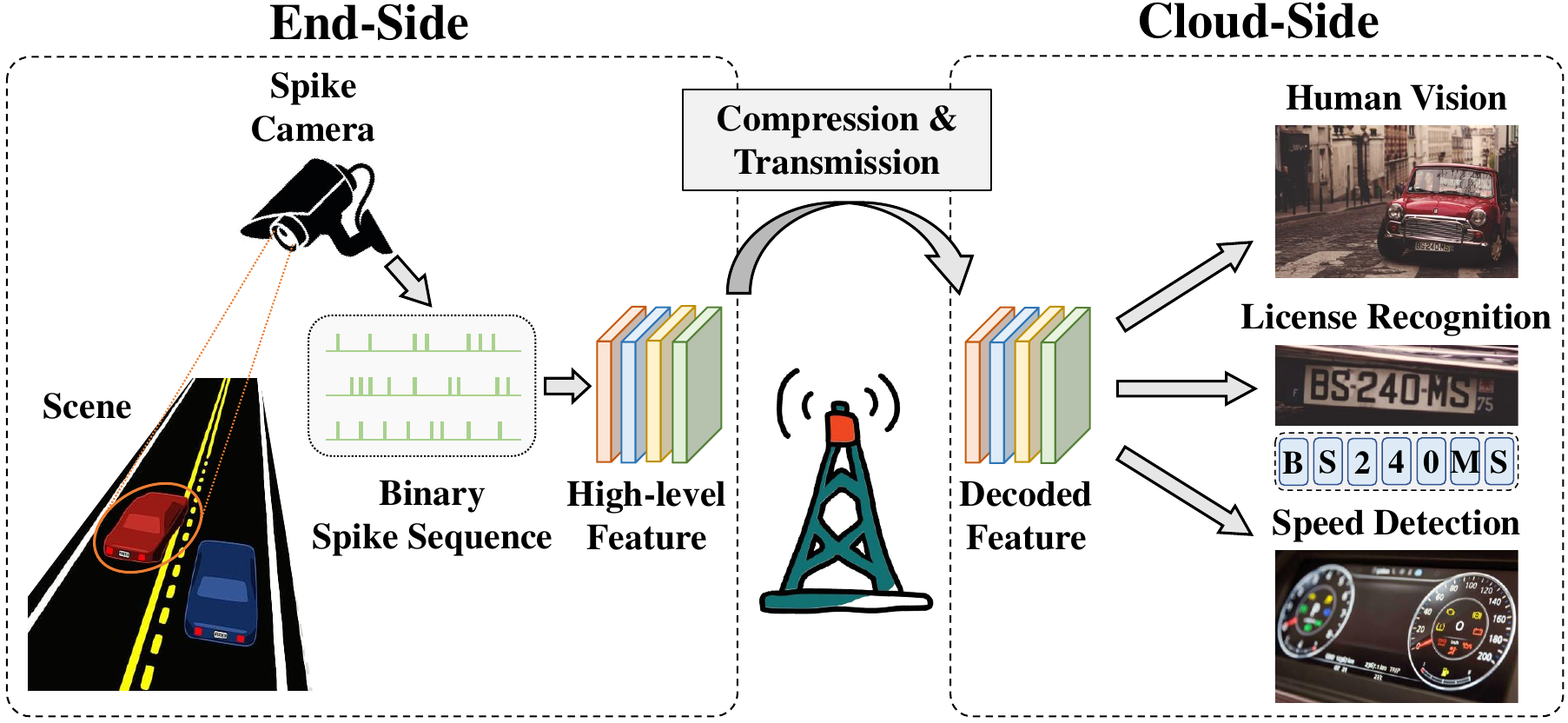}
		\caption{
			Sketch for application of spike vision.
			At the end-side, spike camera captures scenes and generates binary spike sequences, which are extracted as high-level features.
			After compression and transmission, decoded features are utilized for several applications at the cloud-side.
		}
		\label{tisser}
	\end{figure}
	
	Drawing inspiration from the fovea of the mammalian retina~\cite{wassle1991functional}\cite{bringmann2018primate}, a groundbreaking spike camera has been introduced to overcome these limitations~\cite{li2021recent}.
	Spike cameras, by asynchronously recording luminance intensity through spike firing, transcend the constraints of traditional exposure methods.
	This imaging principle affords an ultra-high temporal resolution of 10kFPS, effectively addressing motion blur and content loss in high-speed scenarios.
	Such capabilities enable numerous applications with end-cloud collaboration~\cite{zhang2023application}\cite{rong2021edge}, as depicted in Fig.~\ref{tisser}.
	Scenes are captured and converted into spikes, which are then extracted into features at the end-side. These features undergo compression and transmission, followed by decoding and analysis for various intelligent applications such as scene reconstruction, license recognition, and speed detection.
	However, the ultra-high temporal resolution leads to a significant increase in data volume, exacerbating challenges in storage and transmission.
	For instance, a spike camera with a spatial resolution of 1k$\times$1k and a temporal resolution of 40k generates data at a rate that can impose a significant burden on bandwidth, reaching up to 40Gbps.
	Thus, the development of an efficient compression method for spike data is of paramount importance.
	
	By comprehensively considering the imaging principles of spike cameras and the relationship between spike data and intelligent applications, we introduce a novel compression framework specifically tailored for spike data.
	Our framework comprises three key components: short-term feature extraction, feature compression, and long-term feature analysis.
	Through joint optimization, spike sequences are encoded with minimal bit usage, achieving optimal performance for downstream tasks simultaneously.
	The major contributions of this paper are summarized as follows:
	\begin{itemize}
		\item 
		We conceptualize Spike Coding for Intelligence (SCI), which considers compression efficiency and the performance of spike-based analysis comprehensively.
		We further propose a novel spike compression framework based on the \textit{compress-and-analyze-simultaneously} (\textit{CAAS}) paradigm.
		By end-to-end optimization, our scheme realizes state-of-the-art (SOTA) performance in both compression and intelligent analysis tasks, establishing a new direction for collaborative spike intelligence.
		\item 
		Drawing inspiration from mechanisms in mammalian vision, we propose a dual-pathway architecture to handle spatial semantics and motion information as distinct streams, which are then integrated through a Pathway Fusing Unit (PFU) to address challenges arising in both low and high-speed scenarios effectively.
		To preserve the content fidelity between decoded features and motion vectors, a Feature-level Motion Vector Refinement (FMVR) is introduced to constrain consistency.
		We further propose an Associated Feature Regression (AFR) approach incorporating different motion dynamics in the temporal domain, jointly leveraging the advances of warping and deformation.
		Ablation studies indicate that all technological innovations we proposed yield good results.
		\item 
		Extensive experiments demonstrate the superiority of our method over other approaches, achieving leading performance for intelligent applications at both pixel-level and semantic-level.
		For spike coding, our scheme rebuilds terrific scenes with transmitting minimal bits, reaching a \textbf{17.25\%} BD-rate reduction and obtaining superior subjective quality.
		For spike classification, our method accomplishes high accuracy for datasets S-MNIST, S-CIFAR, and S-CALTECH, exceeding the SOTA method by \textbf{0.5\%}, \textbf{2.5\%}, and \textbf{4.3\%} respectively.
	\end{itemize}
	
	\section{Related Work}
	We comprehensively review related work on spike processing, including spike analysis and compression.
	There is a growing trend of spike-based intelligent applications becoming increasingly popular due to their excellent performance, making spike compression particularly important.
	
	\textbf{Spike Analysis.}
	This includes pixel-level scene reconstruction and semantic-level classification.
	For reconstruction, the scene is estimated from the average spike firing rate within sliding windows, which can hardly handle the rapid change of light intensity.
	To cope with this, Zhao \etal proposed a hierarchical structure to rebuild scenes progressively~\cite{zhao2021spk2imgnet}.
	Based on multi-stage motion estimation and temporal fusion, artifacts and distortion are effectively decreased.
	Furthermore, the structure of the mammalian retina was mimicked for processing visual signals, though this proved challenging.
	For classification, spatial information is supervised by motion characteristics, resulting in better accuracy.
	Binary modulation is introduced to reduce computational consumption, diminishing complexity while maintaining performance~\cite{zhao2023spireco}.
	To further utilize the spatio-temporal correlation within spikes, a transformer-like architecture was proposed for optimizing global association~\cite{yao2023spike}.
	
	\textbf{Spike Compression.}
	Simulating conventional codecs, which are based on partition and prediction, spike sequences were divided into several overlapped voxels~\cite{dong2018spike} and compressed respectively.
	To utilize the correlation between spikes effectively, inter-voxel reference was introduced to achieve better performance.
	Breaking the limitation of binary data format, spikes with discrete values were converted to continuous inter-spike intervals, providing a prerequisite for NN-based compression.
	Feng \etal proposed an end-to-end learnable framework~\cite{feng2023spikecodec}, which decouples the spike compression problem into scene construction plus scene coding tasks, achieving great performance in information fidelity.
	Dong \etal addressed the challenges of lossless compression for continuous spike streams and introduced a learned compression model~\cite{dong2024learned}, achieving state-of-the-art performance.
	
	\textbf{Neural Video Compression.}
	With the emergence of deep learning, neural video compression (NVC) methods have evolved rapidly, leading to significant advancements in the field~\cite{duan2023learned}\cite{jia2022fpx}\cite{ma2019image}.
	DVC~\cite{lu2019dvc} pioneered the first fully neural video compression framework, which replaced traditional optical flow estimation and residual coding modules with trainable neural networks optimized through a rate-distortion loss function.
	Subsequently, DCVC~\cite{li2021deep} leveraged feature-domain contexts as conditions, enabling both encoder and decoder to access rich information for reconstructing high-frequency content, thereby enhancing video quality.
	Building upon DCVC, DCVC-DC~\cite{li2023neural} introduced hierarchical quality patterns across frames to enrich long-term, high-quality temporal contexts and employs a group-based offset diversity mechanism for improved temporal context mining. Additionally, a quadtree-based partitioning strategy is utilized to enhance spatial context diversity during latent representation encoding.
	
	\section{Problem Formulation}
	\subsection{Principles of Spike Generation}
	The spike camera is composed of two-dimensional photosensitive array, which simulates the fovea structure of retina.
	Different from event-based cameras, the spike camera accumulates photons at each position and generates spikes through an integrate-and-fire ($\mathcal{IF}$) process.
	This can be formulated as
	\begin{equation}
		S_t, v_t = \mathcal{IF}(v_{t-1}, I_t),
		\label{if}
	\end{equation}
	where $v_{t-1}$ and $v_t$ represent membrane potential at moment $t-1$ and $t$, while $I_t$ and $S_t$ indicate luminance and spike at moment $t$ respectively.
	By iterating Eq.~\ref{if}, the spike sequence $\{S_t\}$ is generated as following
	\begin{equation}
		\{S_t\} = \mathcal{IF}[v_0, \{I_t\}],
		\label{if_iter}
	\end{equation}
	where $v_0$ represents the initial membrane potential and $\{I_t\}$ demotes the luminance sequence.
	As a matter of fact, pixels in brighter environment can reach threshold faster and thus can fire spikes more frequently.
	This observation aligns with the biological perspective.
	
	The $\mathcal{IF}$ process motioned above can be separated into three steps.
	The intensity of illumination $I_t$ integrates on membrane potential $v_t$, subsequently determining the spike $s_t$ based on whether $v_t$ surpasses threshold $\theta$.
	Finally, $v_t$ undergoes a reset according to $s_t$.
	These three steps can be formalized as
	\begin{equation}\label{integrate}
		v_t = v_{t-1} + \alpha I_t,
	\end{equation}
	\begin{equation}\label{fire}
		S_t = \left\{
		\begin{aligned}
			1,&\quad v_t \ge \theta \\
			0,&\quad otherwise
		\end{aligned}
		\right.,
	\end{equation}
	\begin{equation}\label{reset}
		v_t = \left\{
		\begin{aligned}
			(1 - S_t)v_t,&\quad hard\ scheme \\
			v_t-S_t\theta,&\quad soft\ scheme
		\end{aligned}
		\right.,
	\end{equation}
	where $\alpha$ represents photoelectric conversion efficiency.
	
	We further provide a detailed comparison of the similarities and differences between event and spike camera in terms of imaging principles and data characteristics, as shown in Table~\ref{event-camera}.
	\begin{table*}
		\centering
		\caption{
			Similarities and differences between event and spike camera.
		}	
		\renewcommand{\arraystretch}{1.5}
		\begin{tabular}{@{}l|l|l|l|l|l@{}}
			\bottomrule[1.5pt]
			& \textbf{Imaging Mechanism} & \textbf{Temporal Resolution} & \textbf{Data Structure} & \textbf{Value Range} & \textbf{Condition for Generation} \\
			\hline
			\textbf{Event Camera} & \multirow{2}{*}{asynchronous imaging} & \multirow{2}{*}{ultra-high} & set of $(x,y,t,p)$ & 1 or -1 & $\ln(I(t+\Delta t)) - \ln(I(t)) = p\theta$ \\
			\cline{1-1} \cline{4-6}
			\textbf{Spike Camera} &  &  & tensor with size of $H\times W\times T$ & 1 or 0 & $\alpha\int_{t}^{t + \Delta t}I(\tau)d\tau = \theta$ \\
			\toprule[1.5pt]
		\end{tabular}
		\label{event-camera}
	\end{table*}
	
	\subsection{Theory Conceptualization}
	Differing from images and videos, which are visually interpretable for humans, spike data encapsulates scenes in an abstract manner and is not inherently designed for visual perception.
	This data format offers the advantage of enabling intelligent applications to be executed with high precision and low latency.
	Consequently, the primary objective in decoding spike data is to optimize performance for specific tasks, rather than striving for spike-level fidelity.
	To illustrate this characteristic, we introduce the concept of \textbf{S}pike \textbf{C}oding for \textbf{I}ntelligence (\textbf{SCI}), which comprehensively considers compression efficiency and task performance.
	
	Given task set $\{T_i\}$, $Y_i$ and $\hat{Y_i}$ denote the ground truth and generated result from spike sequence $\{S_t\}$ for task $T_i$.
	The goal of SCI is decreasing the code-length of bit stream for transmission while keeping fidelity between each $Y_i$ and $\hat{Y_i}$.
	The loss function can be formulated as
	\begin{equation}
		\mathcal{L} = \mathcal{R}(\{S_t\}) + \sum_{i}\lambda_i\mathcal{D}_i(Y_i, \hat{Y}_i),
	\end{equation}
	where $\mathcal{R}$ denotes bit-rate estimation, $\mathcal{D}_i$ and $\lambda_i$ denotes distortion and weight for $T_i$.
	
	\begin{figure}
		\centering
		\includegraphics[width=\linewidth]{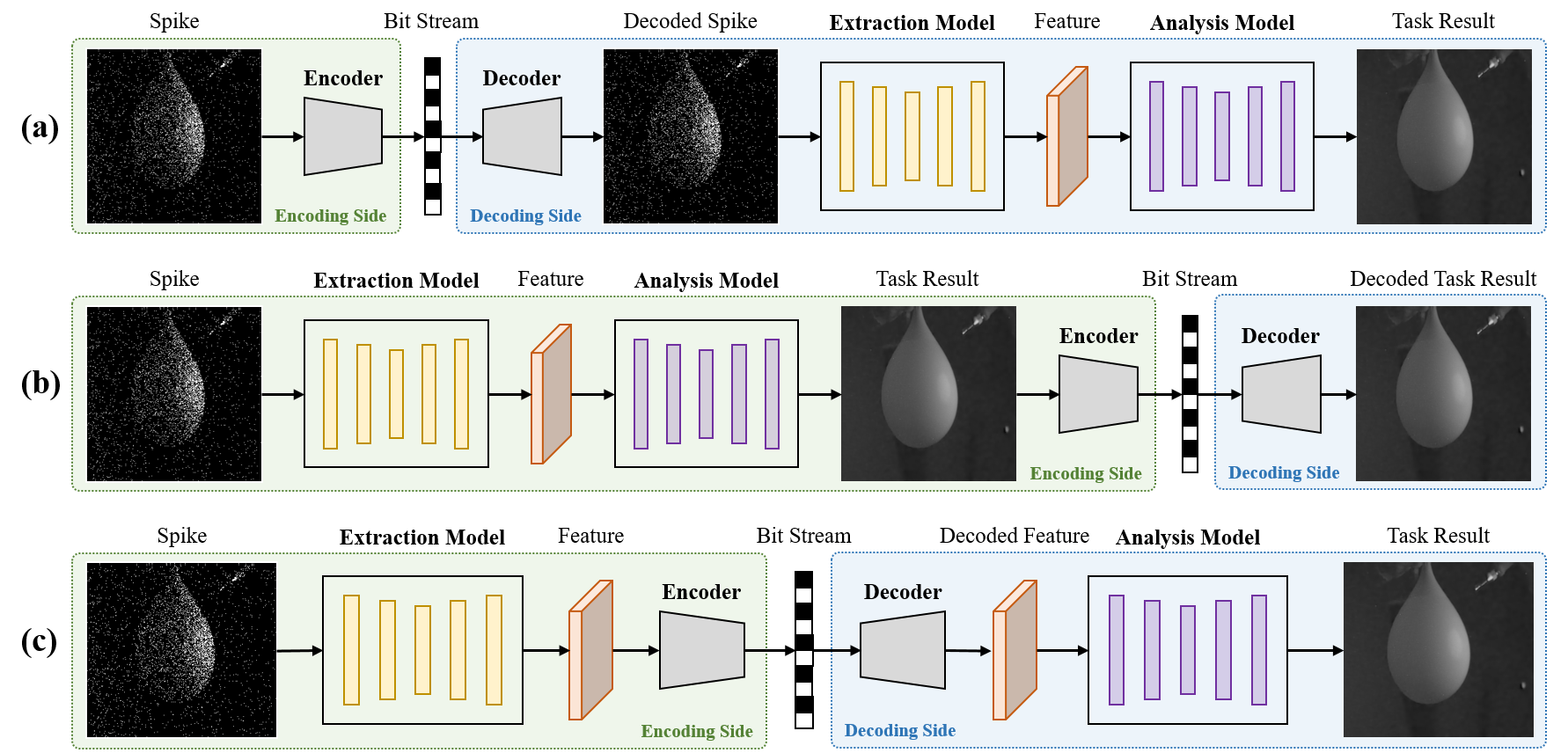}
		\caption{
			Comparison between three paradigms for spike compression and analysis, including 
			(\textbf{a}) \textit{compress-then-analyze} (\textit{CTA}), (\textbf{b}) \textit{analyze-then-compress} (\textit{ATC}) and (\textbf{c}) \textit{compress-and-analyze-simultaneously} (\textit{CAAS}) paradigms.
			Extensive experimental results show \textit{CAAS} paradigm exceeds other two paradigms, providing a novel perspective for SCI.
		}
		\label{paradigm}
	\end{figure}
	\subsection{Paradigm Comparison}
	Referring to task-oriented compression frameworks~\cite{yang2021video}\cite{fischer2020video}, we introduce three SCI paradigms as shown in Fig.~\ref{paradigm}.
	Green box denotes the encoding side, while blue represents the decoding side.
	For the \textit{compress-then-analyze} (\textit{CTA}) paradigm depicted in Fig.~\ref{paradigm}(a), spikes are first encoded into a bitstream and then decoded back to spikes, which are subsequently analyzed for downstream tasks.
	However, the extremely weak spatio-temporal continuity leads to poor efficiency in compression and noticeable distortion in reconstruction, which further results in inferior task performance.
	For the \textit{analyze-then-compress} (\textit{ATC}) paradigm depicted in Fig.~\ref{paradigm}(b), spikes are first analyzed for downstream tasks, and the analysis results are then compressed.
	Nevertheless, the results for the task may be unfriendly for compression, and the bitstream is not versatile enough for multiple tasks.
	
	Considering compression efficiency, task performance, and bitstream versatility, we propose the \textit{compress-and-analyze-simultaneously} (\textit{CAAS}) paradigm depicted in Fig.~\ref{paradigm}(c).
	Spikes are first extracted into features, which are then compressed and analyzed for downstream tasks.
	On the one hand, the strong spatio-temporal correlation within features can be utilized to reduce the bit rate.
	On the other hand, features aggregate low-level semantic information, which can be repurposed in several high-level tasks.
	Extensive experiments demonstrate the superiority of the \textit{CAAS} paradigm over the \textit{CTA} and \textit{ATC} paradigms, providing a novel perspective for SCI.
	
	Given the limited computational capability at the end-side, which severely restricts the usage of spike cameras, our proposed \textit{CAAS} paradigm is resource-efficient, making it more applicable to the end-cloud collaborative architecture.
	The electrical power limitation at the end-side ($<$10W) restricts its computational capability ($<$10GFLOPS), necessitating a simplification of the encoding procedure to reduce delay and avoid overheating.
	In contrast, the cloud-side has abundant computing resources ($>$100TFLOPS) and can offload complex computational operations.
	As such, we design the framework with lower complexity on the end side and higher complexity on the cloud side, achieving a balance between encoding and decoding based on the difference in end-cloud computing resources.
	The spike camera, serving as the end-side (edge device), converts spikes to features through an extraction model and then compresses them into a bitstream via an encoding model.
	The cloud-side transforms the bitstream into features through a decoding model and applies them for intelligent applications by analysis models.
	
	\section{Methodology}
\begin{figure*}
		\centering
		\includegraphics[width=\linewidth]{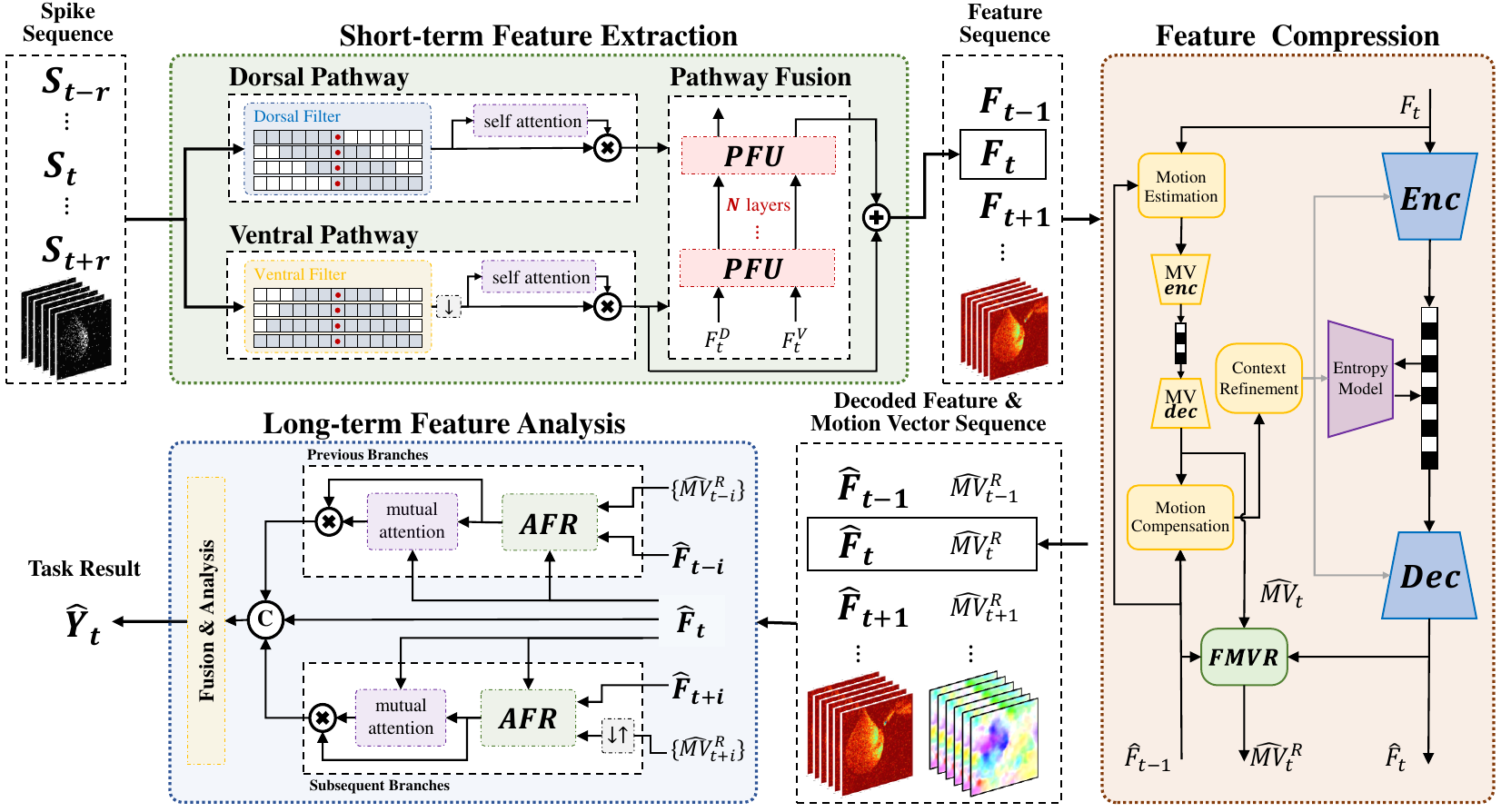}
		\caption{
			Framework of method which compresses spike sequence and analyzes for downstream tasks simultaneously.
			The whole process is mainly separated into three modules.
			The short-term feature extraction module converts spike sequence into compression-friendly feature ($F_t\in [0,1]^{h\times w}$).
			The feature compression module effectively encodes feature sequence, eliminating intra- and inter-feature redundancies.
			The long-term feature analysis module optimizes for downstream task performance using decoded feature sequence and refined motion vector sequence.
		}
		\label{framework}
	\end{figure*}

	The overarching framework of our study is illustrated in Fig.~\ref{framework}, outlining a three-step process for the efficient compression of spike sequences and their subsequent analysis in downstream applications.
	Initially, the spike sequence $\{S_t\}$ is transformed into a feature sequence $\{F_t\}$, which is amenable to coding.
	Subsequently, $\{F_t\}$ undergoes effective compression, achieved by reducing both intra- and inter-feature redundancies.
	In the final step, we utilize the decoded feature sequence $\{\hat{F}_t\}$ along with the refined motion vector sequence $\{\hat{MV}_t^{R}\}$ to enhance performance in downstream tasks.
	Detailed explanations of our methodology are provided in the following subsections.
	
	\begin{figure}
		\centering
		\includegraphics[width=1\linewidth]{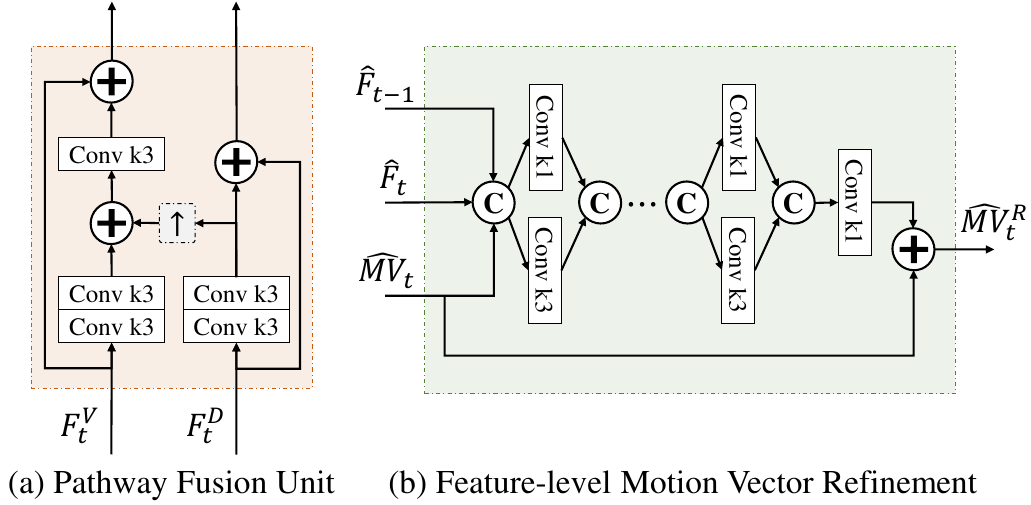}
		\caption{
			Detailed structure of (\textbf{a}) PFU and (\textbf{b}) FMVR.
			The PFU fuses spatial semantic information from dorsal pathway and motion characteristic from ventral pathway to generate features which are compression-friendly.
			The FMVR utilizes decoded features to constrains consistency of content between feature and latent domain.
		}
		\label{pfu}
	\end{figure}
	
	\begin{figure*}
		\centering
		\includegraphics[width=\linewidth]{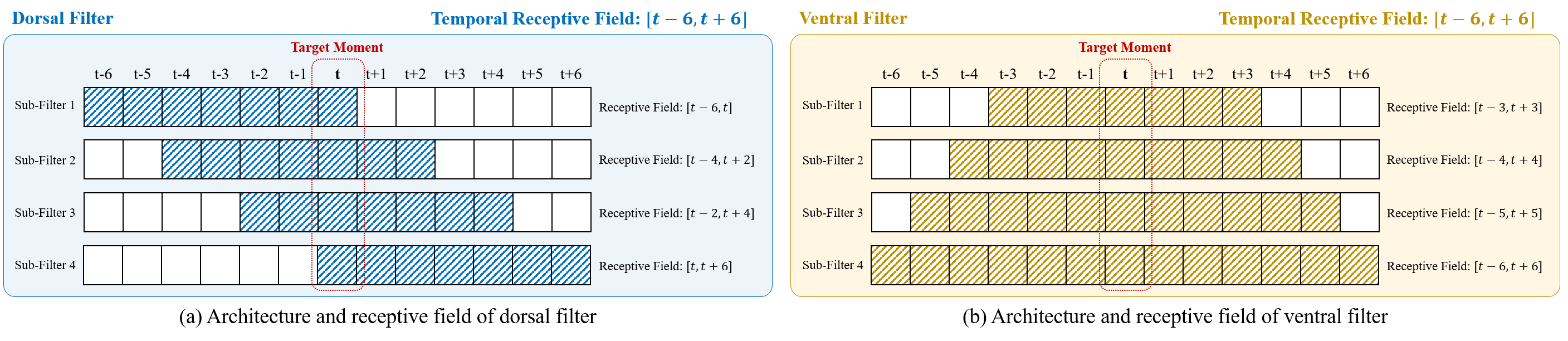}
		\caption{
			Comparison of architecture and temporal receptive field between dual-pathway filters.
			The dorsal filter consists of sliding window-based temporal sub-filters, whereas the ventral filter comprises multi-scale temporal sub-filters.
		}
		\label{filter}
	\end{figure*}
	\subsection{Short-term Feature Extraction}
	This module leverages the correlation in spike sequences $\{S_t\}$ to amalgamate spatio-temporal data and formulate an efficient feature representation $F_t$. Inspired by the Two-streams Hypothesis\cite{kravitz2011new}\cite{milner2008two}, which postulates the bifurcation of visual information into two distinct streams, we develop a dual-pathway architecture.
	The dorsal pathway focuses on motion characteristics, detecting positions and changes with the assistance of a sliding window-based temporal filter as shown in Fig.~\ref{filter} (a).
	Conversely, the ventral pathway integrates spatial semantics and perceives structural information via a multi-scale temporal filter as shown in Fig.~\ref{filter} (b).
	Both filters are binarized and enhanced by a self-attention mechanism sub-module~\cite{vaswani2017attention}.
	
	Emerging biological research indicates a symbiotic relationship between the dorsal and ventral streams, especially in processing detailed semantics~\cite{van2015interactions}\cite{ayzenberg2023temporal}.
	This observation prompts the creation of the Pathway Fusion Unit (PFU), which is depicted in Fig.~\ref{pfu}(a). By employing $N$ layers of PFUs, features from dorsal and ventral pathways are effectively collaborated, producing a feature $F_t$ enriched with integrated spatio-temporal information and friendly for compression, as formulated below:
	\begin{equation}
		F_t = \mathcal{PFU}^{[N]}(F_t^D, F_t^V) + F_t^V,
	\end{equation}
	where $F_t^D$ and $F_t^V$ denote features derived from the dorsal and ventral pathways, respectively.
	
	\subsection{Feature Compression}
	This module employs advanced compression techniques to encode the feature sequence $\{F_t\}$ into a compact bitstream while ensuring that the decoded feature sequence $\{\hat{F}_t\}$ has minimal distortion.
	The architectural design closely resembles that of SOTA learning-based video codecs~\cite{lu2019dvc}\cite{li2021deep}\cite{li2023neural}.
	Initially, $F_t$ and the preceding decoded feature $\hat{F}_{t-1}$ are used to estimate the decoded motion vector $\hat{MV}_t$.
	Based on $\hat{F}_{t-1}$ and $\hat{MV}_t$, contextual conditions for the encoder and decoder are established.
	$F_t$ then undergoes encoding into a bitstream in the most compact form, which is decoded as $\hat{F}_t$ with minimal distortion.
	
	In conventional feature codecs, $\hat{MV}_t$ is only used for decoding $\hat{F}_t$ and will eventually become deprecated.
	We recognize that $\hat{MV}_t$ encapsulates motion details between $\hat{F}_{t-1}$ and $\hat{F}_t$, which can be invaluable for subsequent alignment.
	However, $\hat{MV}_t$ only works in the latent domain, which is incompatible with feature contents~\cite{skorokhodov2021aligning}.
	Thus, we propose the Feature-level Motion Vector Refinement (FMVR), which is shown in Fig.~\ref{pfu}(b).
	Composed of convolution layers with small kernels, FMVR focuses on local similarity and constrains content consistency between the feature and latent domains~\cite{1991small}.
	\begin{equation}
		\hat{MV}_t^R = \mathcal{FMVR}(\hat{MV}_t|\hat{F}_{t-1}, \hat{F}_t).
	\end{equation}
	$\hat{MV}_t^R$ is then preserved to fully leverage motion information for alignment.

	\begin{figure}
		\centering
		\includegraphics[width=1.05\linewidth]{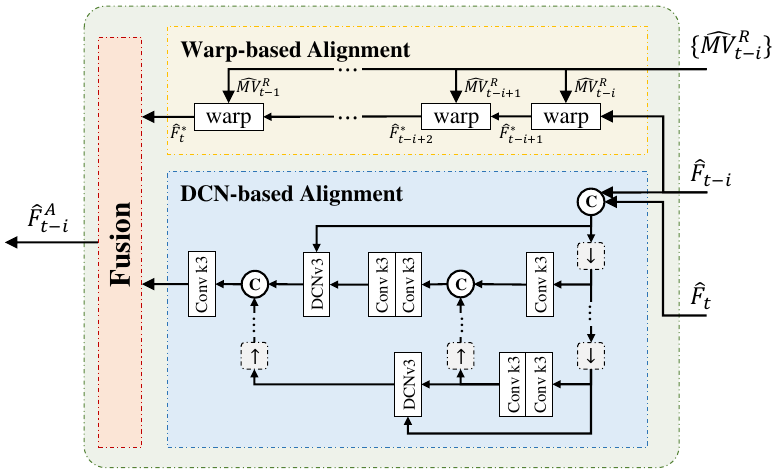}
		\caption{
			Detailed structure of AFR, which incorporates different motion dynamics in temporal domain.
			Precise regressed feature is generated via leveraging the advances of warping and deformation approaches, which exhibits smoother transitions and leads to more accurate result for tasks.
		}
		\label{fau}
	\end{figure}
	\subsection{Long-term Feature Analysis}
	This module regresses $\{\hat{F}_t\}$ to moment $t$ with the assistance of $\{\hat{MV}_t^R\}$ and fuses them to generate the result $\hat{Y}_t$ for the task.
	By aggregating abundant context, artifacts are effectively eliminated, and performance is significantly improved.
	$\{\hat{F}_t\}$ and $\{\hat{MV}_t^R\}$ are split into previous and subsequent branches, within which features are aligned separately.
	Traditional alignment methods predominantly rely on optical flow and assume that each pixel moves from only one previous location~\cite{alldieck2017optical}\cite{brox2010large}, exhibiting limited diversity and imprecision at a fine-grained level.
	Recent advancements have leveraged the Deformable Convolution Network (DCN)~\cite{zhu2019deformable}\cite{dai2017deformable} to produce multiple groups of offsets and masks, enhancing sub-pixel level precision and fault tolerance~\cite{wang2019edvr}\cite{huang2022improved}.
	Nonetheless, DCN exhibits training instability, and its spatial receptive field is constrained in capturing large-scale motion~\cite{chan2021understanding}.
	Hence, we amalgamate both warping and deformation approaches, introducing the Associated Feature Regression (AFR) depicted in Fig.~\ref{fau}.
	\begin{equation}
		\hat{F}_{t-i}^A = \mathcal{AFR}(\hat{F}_t,\hat{F}_{t-i},\{\hat{MV}_{t-i}^R\}),
	\end{equation}
	where $\hat{F}_{t-i}^A$ represents the aligned feature at moment $t$.
	Large-scale and small-scale motions are comprehensively captured, resulting in more precise alignment and superior fault tolerance.
	After alignment, preliminary results are generated and then fused to produce $\hat{Y}_t$.
	Experimental results demonstrate that our AFR outperforms both warping and deformation methods in feature regression.

	\section{Experimental Results}
	\subsection{Settings}
	Tasks encompassing various levels are utilized to assess the versatility of the decoded feature sequence $\{\hat{F}_t\}$, specifically focusing on \textbf{scene reconstruction} and \textbf{classification} tasks.
	In the context of scene reconstruction, rich structural and textural details are acquired to ensure high pixel-level fidelity~\cite{starikov2007input}, which is evaluated on artificial \textit{S-VIMEO}.
	In contrast, classification highlights the capability to extract high-level semantic information~\cite{lu2007survey}.
	This is assessed on \textit{S-MNIST}, \textit{S-CIFAR}, and \textit{S-CALTECH} referred to~\cite{zhao2023spireco}.
	The ventral filter consists of 4 sub-filters with temporal receptive fields (TRFs) spreading from center towards both ends (TRF=7/9/11/13).
	The dorsal filter consists of 4 sub-filters with a fixed window size sliding backward (window\_size=10).
	Extensive experimental results showcase the exceptional performance of our model across both tasks.
	
	\begin{figure}
		\centering
		\includegraphics[width=\linewidth]{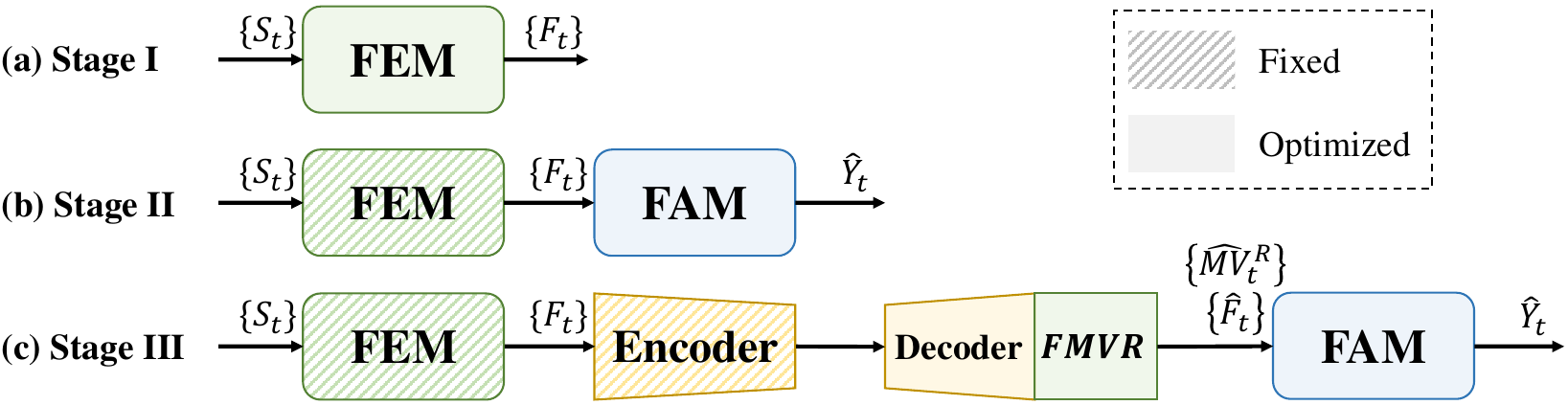}
		\caption{
			Sketch for multi-stage training strategy.
			(\textbf{a}) Stage I: train the Feature Extraction Module.
			(\textbf{b}) Stage II: fix the Feature Extraction Module and train the Feature Analysis Module.
			(\textbf{c}) Stage III: fix the Feature Extraction Module and encoder and train the decoder, FMVR and Feature Analysis Module.
		}
		\label{train_strategy}
	\end{figure}
	\begin{figure}
		\centering
		\includegraphics[width=\linewidth]{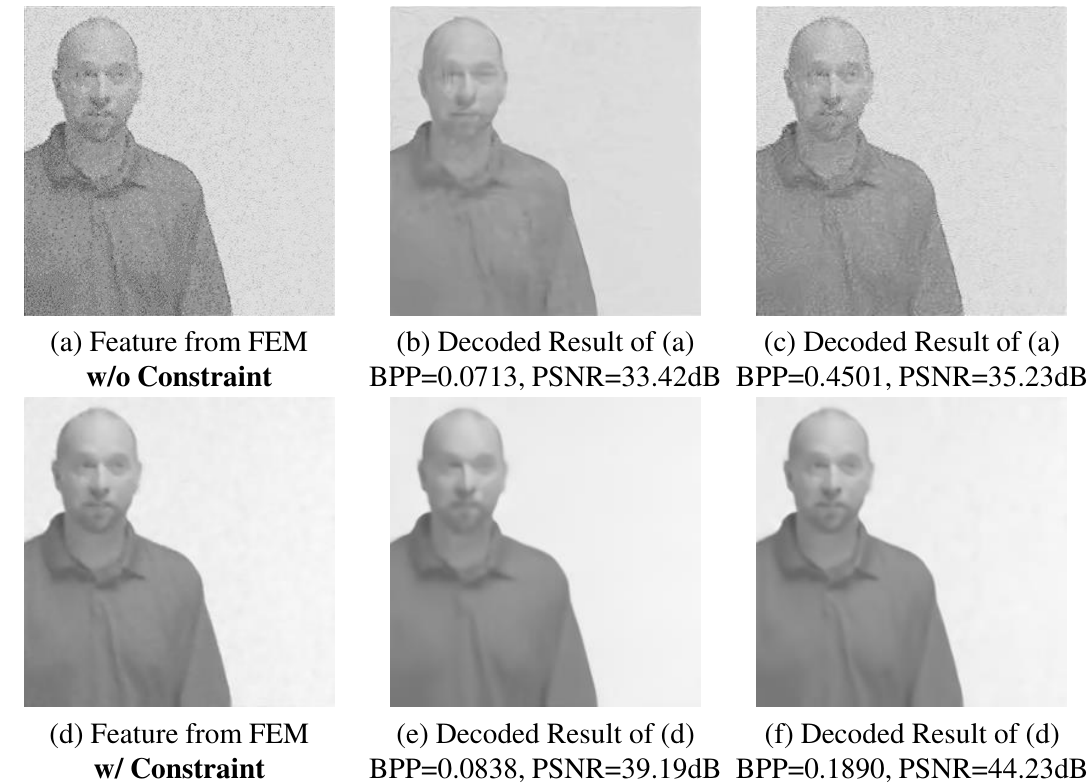}
		\caption{
			Visual comparison between of the feature extracted from the spike sequence with and without constraint.
			(\textbf{a}) feature extracted from the Feature Extraction Module without constraint.
			(\textbf{b$\sim$c}) decoded results of (a) with different quantization steps.
			(\textbf{d}) feature extracted from the Feature Extraction Module with constraint.
			(\textbf{e$\sim$f}) decoded results of (d) with different quantization steps.
			Results show that the smoothness of feature plays a pivotal role in determining the codec's efficiency.
			Zoom in for better visibility.
		}
		\label{constraint}
	\end{figure}
	
	\subsection{Training Strategy}
	The complexity of our model, with its numerous modules, poses a challenge for a unified training approach, as this might impede convergence and lead to poor performance.
	To address this, we advocate for a multi-stage training strategy, illustrating in Fig.~\ref{train_strategy}.
	In the initial phase, we focus solely on training the Feature Extraction Module (FEM), as shown in Fig.~\ref{train_strategy}(a).
	The objective here is to regulate the network's output to ensure the smoothness of the feature $F_t$, the importance of which will be elucidated subsequently.
	During the second phase, the FEM is fixed, and we train distinct Feature Analysis Modules (FAMs) for various tasks, as shown in Fig.~\ref{train_strategy}(b).
	These modules are regressed using either the Mean Squared Error (MSE) loss for scene reconstruction or Cross-Entropy (CE) loss for classification to attain optimal performance.
	In the final stage, pre-trained codecs are integrated between the FEM and FAMs.
	While keeping the FEM and encoder fixed, we only fine-tune the decoder, FMVR and FAMs, as shown in Fig.~\ref{train_strategy}(c).
	Empirical results demonstrate that this structured training approach not only makes the compressed bitstream suitable for various downstream applications but also significantly enhances performance across different tasks.
	
\begin{figure}
		\centering
		\includegraphics[width=.9\linewidth]{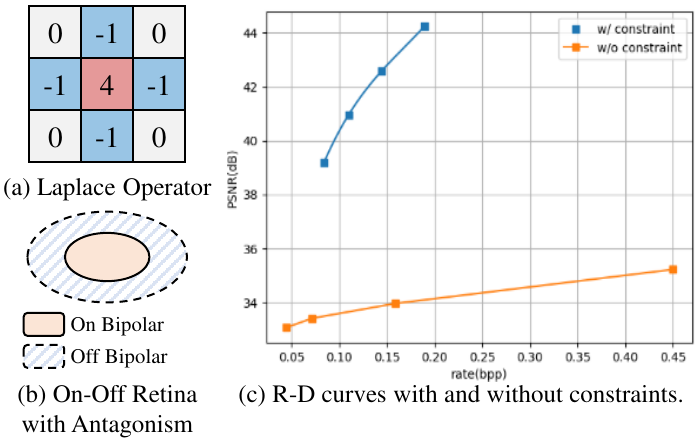}
		\caption{
			(\textbf{a}) Sketch for the Laplacian operator.
			(\textbf{b}) Sketch for the On-Off retina with antagonism.
			The substantial similarity observed between these two structures validates the rationale behind employing Laplacian operator constraints to ensure feature smoothness.
			(\textbf{c}) Rate-distortion curve for methods with and without constraint.
		}
		\label{laplace_operator}
	\end{figure}

	Although the intermediate feature $F_t$ does not exert a direct influence on the loss function, its degree of smoothness plays a pivotal role in determining the codec's efficiency. This further exerts a substantial influence on both the bit-rate and the efficacy of downstream intelligent tasks. Fig.~\ref{constraint} illustrates this concept by visualizing the features extracted from the spike sequence, thereby underscoring the criticality of imposing constraints on $F_t$.
	BPP and PSNR are acronyms for Bit Per Pixel and Peak Signal-to-Noise Ratio.
	In Fig.~\ref{constraint}(a), where $F_t$ remains unconstrained, the resultant high-frequency noise pervading the background not only fails to contribute positively to the final task performance but also detrimentally affects the codec's functionality. Fig.~\ref{constraint}(b$\sim$c) demonstrates that in the absence of constraints, the codec escalates the bit rate ineffectively, primarily to restore high-frequency noise. In stark contrast, Fig.~\ref{constraint}(d) reveals that constraining $F_t$ leads to a more uniform background, facilitating easier compression. Such a constraint enables the codec to prioritize the restoration of content-related high-frequency textures, which, as the bit rate escalates, results in an enhanced recovery of details, as vividly depicted in Fig.~\ref{constraint}(e$\sim$f).
	Additionally, we have graphed the Rate-Distortion curves under both constrained and unconstrained scenarios, as depicted in the Fig.~\ref{laplace_operator}(c). The results reveal that, under constraints, the encoder exhibits superior bit-efficiency and reconstruction quality. This further corroborates the notion that smoother features are more amenable to compression. These findings highlight the effectiveness of applying constraints in enhancing the overall performance of the encoding process, particularly in terms of balancing the trade-off between bit rate and the fidelity of decoding.
	
	\begin{table}
		\centering
		\small
		\renewcommand\arraystretch{1.3}
		\begin{tabular}{l|c|c}
			\bottomrule
			& \textbf{w/o constraint} & \textbf{w/ constraint} \\ \hline \hline
			L1-Norm of $\mathbb{OP}_L$ & 0.1079         & 0.0090        \\ \hline
			L2-Norm of $\mathbb{OP}_L$ & 0.0341         & 0.0003       \\
			\bottomrule
		\end{tabular}
		\vspace{5pt}
		\caption{
			L1- and L2-Norms of the Laplace operation.
			Results substantiate the Laplacian operator's effectiveness in enhancing smoothness, and reaffirm its utility in refining the feature representation for more efficient compression.
		}
		\label{laplace}
	\end{table}
	
	For further evaluation, we employed the L1- and L2-Norms of the Laplacian operator ($\mathbb{OP}_L$) to quantify the smoothness of $F_t$, shown in Fig.~\ref{laplace_operator}.
	In the realm of image processing, the Laplacian operator is utilized to compute the second-order derivatives of pixel in both horizontal and vertical directions, facilitating the extraction of high-frequency content such as edge textures (Fig.~\ref{laplace_operator}(a)).
	This approach mirrors processes observed in biological vision systems, where ganglion cells, while integrating visual signals from bipolar cells, introduce an antagonistic mechanism to reduce redundancy in the optic nerve (Fig.~\ref{laplace_operator}(b)).
	This antagonism is typically represented by the Difference of Gaussian receptive fields (DoGs), akin to an isotropic Laplacian operator in its distribution.
	This analogy underscores that employing Laplacian operator-based constraints can effectively eliminate redundant high-frequency noise, thereby enhancing the compressibility of $F_t$.
	As indicated in Table~\ref{laplace}, the L1- and L2-Norms of the Laplacian operation for the constrained features exhibit a significant reduction. This not only substantiates the Laplacian operator's effectiveness in enhancing smoothness but also reaffirms its utility in refining the feature representation for more efficient compression.

	\subsection{Reconstruction Oriented Compression}
	
	\begin{figure*}
		\centering
		\includegraphics[width=\linewidth]{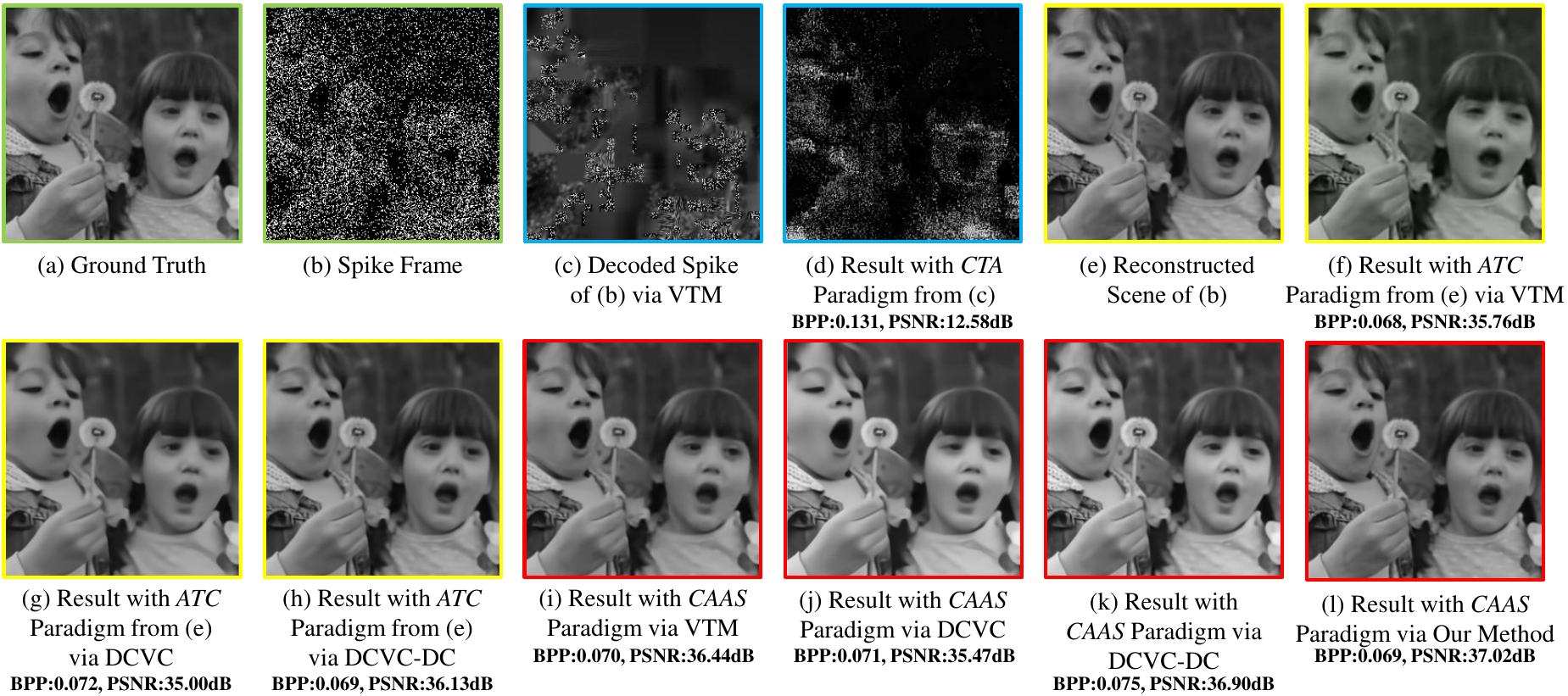}
		\caption{
			Visualization for \textbf{(a)} ground truth, \textbf{(b)} spike frame, \textbf{(c)} decoded spike frame from (b) via VTM, \textbf{(d)} reconstructed scene from (c), \textbf{(e)} reconstructed scene from (b), \textbf{(f$ \sim $h)} decoded scene from (e) via VTM, DCVC and DCVC-DC, \textbf{(i$ \sim $l)} decoded and reconstructed scene from (b) via VTM, DCVC, DCVC-DC and our method.
			Subjective results show our method outperforms other approaches with \textit{CTA}, \textit{ATC} and \textit{CAAS} paradigms, establishing a strong baseline.
			Zoom in for better visibility.
		}
		\label{subjective}
	\end{figure*}
	To facilitate a comprehensive comparison of subjective quality, we depict the reconstructed scenes using different paradigms via various codecs, as shown in Fig.~\ref{subjective}.
	For the \textit{CTA} paradigm, the limited spatio-temporal continuity within spike sequences leads to pronounced blocking artifacts in the reconstruction, resulting in suboptimal reconstructed outcomes.
	For the \textit{ATC} paradigm, distortions present in the reconstructed result from raw spikes are further accumulated and magnified with subsequent compression.
	In contrast, the \textit{CAAS} paradigm enables approaches to simultaneously achieve minimal bit rates and distortion from raw spikes, resulting in superior performance compared to other methods.
	
\begin{figure}
		\centering
		\includegraphics[width=.9\linewidth]{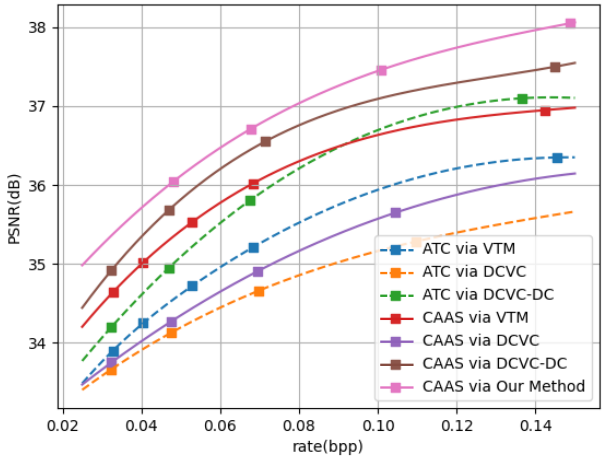}
		\caption{
			Rate-distortion curve for approaches with \textit{ATC} and \textit{CAAS} paradigm.
			BPP is short for bit-per-pixel.
			Results illustrate that performance of our method exceeds that of all other approaches significantly, achieving 17.25\% BD-rate reduction.
		}
		\label{rd_curve}
	\end{figure}

	\textbf{Rate-distortion loss.}
	To evaluate objective performance, we compute the PSNR between the ground truth and the scene reconstructed with different paradigms via variable codecs.
	The performance of approaches with the \textit{ATC} and \textit{CAAS} paradigms is indicated via Rate-Distortion curves shown in Fig.\ref{rd_curve}.
	It is observed from the curves that our method demonstrates significantly better compression efficiency and reduced reconstruction distortion across all bit rates.
	Furthermore, the results also underscore the superiority of our proposed \textit{CAAS} paradigm.
	Employing the same codec, the \textit{CAAS} paradigm exhibits enhanced performance compared to the \textit{ATC} paradigm, laying the groundwork for future research.
	Specific BD-rate\cite{bjontegaard2001calculation} reductions are presented in Table~\ref{bd-rate}, illustrating that our method outperforms other approaches by at least 17.25
	
	The implications of employing the \textit{CTA} paradigm are detailed in Table~\ref{rd_cta}.
	The underlined values indicate that the compression objectives are unmet, given that the bit rate of raw spike sequences stands at 1 BPP.
	Navigating the simultaneous optimization of bit rate and decoding quality poses a formidable challenge within the \textit{CTA} paradigm, particularly in the context of standard bit-rate ranges.
	This challenge is primarily attributed to the inherent binary characteristic of spike sequences, which results in diminished spatio-temporal continuity and consequently leads to volatile reference relationship dynamics in both spatial and temporal dimensions.
	
	\begin{figure*}
		\centering
		\includegraphics[width=\linewidth]{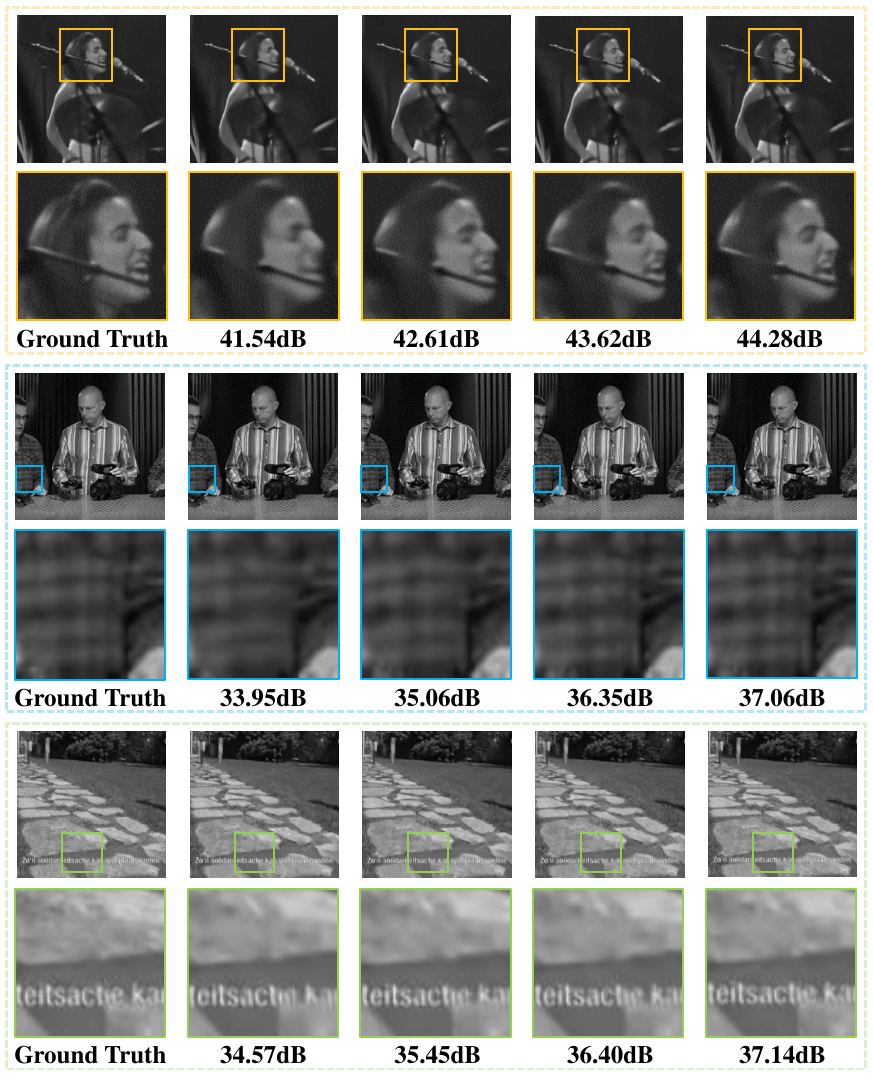}
		\caption{
			Visualization of decoded and reconstructed scenes for simple scenarios.
			The first two rows, middle two rows and last two rows are selected from sequence 00001-0410, 00001-0330 and 00001-0160 respectively.
			The first column is ground truth, while the last four columns are decoded and reconstructed results with average bit-rate of 0.048, 0.068, 0.101, 0.149 BPP.
		}
		\label{recon1}
	\end{figure*}
	
	\begin{table}[]
		\centering
		\small
		\renewcommand\arraystretch{1.3}
		\begin{tabular}{c|c|c|c|l|c|c}
			\bottomrule
			\multicolumn{7}{c}{\textit{CTA} Paradigm via}                                                                                                         \\ 
			\multicolumn{2}{c|}{\textbf{DCVC}}                   & \multicolumn{2}{c|}{\textbf{DCVC-DC}}                & \multicolumn{1}{l}{} & \multicolumn{2}{c}{\textbf{VTM}}                     \\ \hline \hline
			BPP                  & PSNR                 & BPP                  & PSNR                 & QP & BPP                  & PSNR                 \\ \hline
			0.60                 & 13.29                & \underline{1.25}                 & 19.31                & 40 & 0.13                 & 12.2                 \\ \hline
			\underline{1.00}                 & 14.42                & \underline{1.55}                 & 19.43                & 42 & 0.56                 & 16.9                 \\ \hline
			\underline{1.32}                 & 15.04                & \underline{1.83}                 & 19.51                & 44 & 0.63                 & 17.42                \\ \hline
			\underline{1.70}                 & 15.50                & \underline{2.11}                 & 19.69                & 46 & \underline{1.32}                 & 20.36                \\ 
			\bottomrule
		\end{tabular}
		\vspace{5pt}
		\caption{
			Rate-distortion performance for \textit{CTA} paradigm via VTM, DCVC and DCVC-DC.
			The \underline{underline} values indicate the requirement of compression is not accomplished, for bit-rate of raw spike sequences is 1 BPP.
		}
		\label{rd_cta}
	\end{table}
	
\begin{figure}
		\centering
		\includegraphics[width=\linewidth]{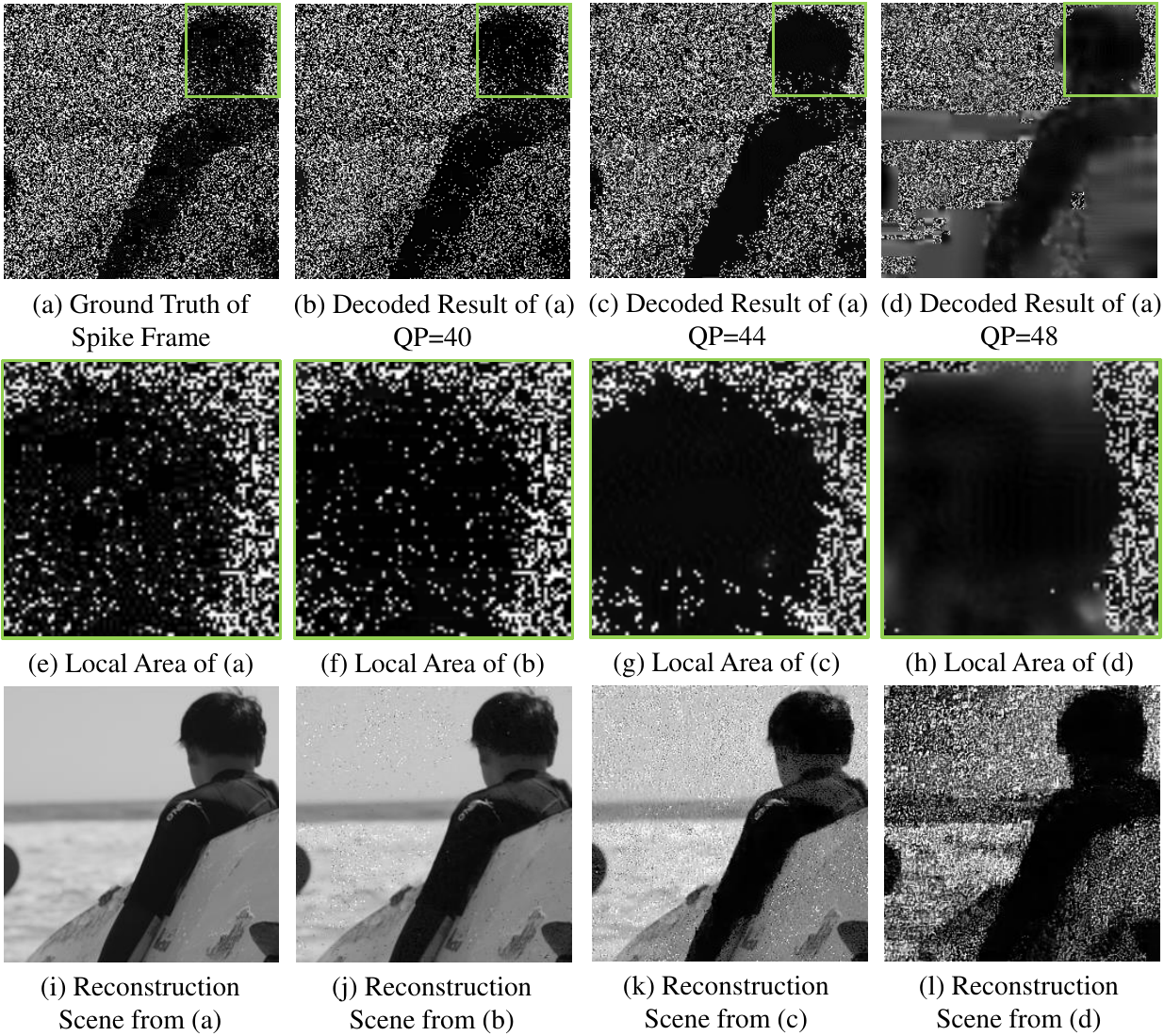}
		\caption{
			Visual comparison of decoded spike frames via VTM with different QP and their reconstructed scenes.
			(\textbf{a}) ground truth of spike frame.
			(\textbf{b$\sim$d}) decoded results via VTM with QP=40, 44 and 48.
			(\textbf{e$\sim$h}) local area of (a$\sim$d).
			(\textbf{i$\sim$l}) reconstruction scenes from (a$\sim$d).
			Results indicate that the inherent binary characteristic of spike sequence culminates in diminished spatio-temporal continuity, leading to severe distortion for compression and downstream intelligent applications.
			Zoom in for better visibility.
		}
		\label{cta_subjective}
	\end{figure}

	To further illuminate this phenomenon, we showcase visualizations of the decoded outcomes utilizing the \textit{CTA} paradigm via VTM, as depicted in Fig.\ref{cta_subjective}.
	It becomes evident, particularly with increasing QP values, that the encoder progressively disregards spikes within areas of sparse population, as evidenced in Fig.\ref{cta_subjective}(f$\sim$g).
	Additionally, pronounced block artifacts are observed in the decoded spike sequences, exemplified in Fig.\ref{cta_subjective}(h), which significantly compromise information fidelity.
	Given the critical reliance of many spike-based intelligent applications on the precise timing of spike occurrences, such high distortion in the decoded spikes substantially hampers the efficacy of downstream tasks.
	The adverse effects of this distortion are further accentuated by the emergence of noise and discontinuities in the reconstructed scenarios, as illustrated in Fig.\ref{cta_subjective}(j$\sim$l).
	
	\begin{table}[]
		\centering
		\small
		\renewcommand\arraystretch{1.3}
		\begin{tabular}{l|c|c|c}
			\bottomrule
			& VTM~\cite{bross2021overview}      & DCVC~\cite{li2021deep}     & DCVC-DC~\cite{li2023neural}  \\ \hline\hline
			\textbf{Ours}    & \textbf{-36.50\%} & \textbf{-63.38\%}         & \textbf{\textcolor{red}{-17.25\%}} \\
			\bottomrule
		\end{tabular}
		\vspace{5pt}
		\caption{
			BD-rate reduction between our method and other approaches with \textit{CAAS} paradigm.
			Negative values represent gain while positive values represent loss.
		}
		\label{bd-rate}
	\end{table}
	
	\textbf{Image quality assessment.}
	We employ non-reference Image Quality Assessment (IQA) metrics to evaluate the subjective quality, aligning with human perception~\cite{ospina2015non}.
	Several IQA metrics, including NIQE~\cite{mittal2012making}, MUSIQ~\cite{ke2021musiq}, and BRISQUE~\cite{mittal2012no}, are selected for comprehensive analysis, with results presented in Table~\ref{iqa}.
	Our method achieves optimal performance across all metrics, demonstrating significant improvements over other approaches in terms of human perception.
	Simultaneously, this underscores the rationality and effectiveness of drawing inspiration from the mammalian vision system.
	
	\begin{table}[]
		\centering
		\small
		\renewcommand\arraystretch{1.3}
		\begin{tabular}{l|c|c|c|c}
			\bottomrule
			& \multicolumn{4}{c}{\textit{CAAS} Paradigm via}                                                       \\
			\multicolumn{1}{c|}{IQA Metric} & \multicolumn{1}{c|}{VTM} & \multicolumn{1}{c|}{DCVC} & \multicolumn{1}{c|}{DCVC-DC} & \textbf{Ours}  \\ \hline\hline
			\textbf{NIQE}\cite{mittal2012making} ($\downarrow$)                            & 9.71                    & 10.18                      & 8.11                         & \textbf{7.71}  \\ \hline
			\textbf{MUSIQ}~\cite{ke2021musiq} ($\uparrow$)                           & 32.58                    & 31.61                     & 39.19                        & \textbf{39.84} \\ \hline
			\textbf{BRISQUE}\cite{mittal2012no} ($\downarrow$)                         & 50.34                    & 53.25                     & 50.32                        & \textbf{49.54} \\
			\bottomrule
		\end{tabular}
		\vspace{5pt}
		\caption{
			The IQA performance comparison between approaches with \textit{CAAS} paradigm via VTM, DCVC, DCVC-DC and our method.
			($\uparrow$) and ($\downarrow$) represent greater value is better and contrary.
			Results indicate our method achieves optimal on all metrics.
		}
		\label{iqa}
	\end{table}
	
	\subsection{Classification Oriented Compression}
\begin{figure}
		\centering
		\includegraphics[width=.9\linewidth]{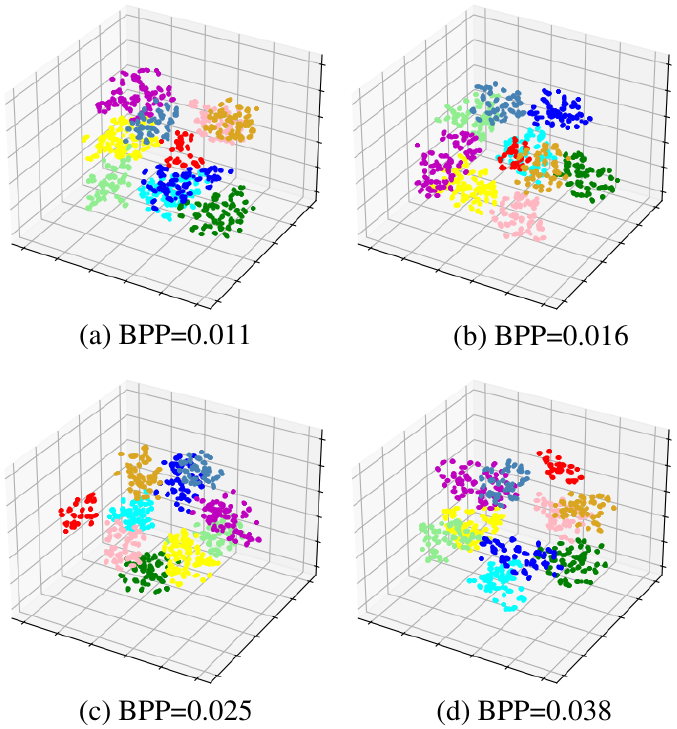}
		\caption{
			Visualization of category aggregation within dataset S-MNIST after compression with various BPP.
			Despite being compressed, results show that samples of the same category are still largely clustered together, leading to high accuracy for classification.
		}
		\label{tsne}
	\end{figure}

	We conduct thorough evaluations across all datasets to affirm the superiority of our methodologies.
	Notably, our method's classification accuracy significantly surpasses that of the SOTA approach~\cite{zhao2023spireco} when compression is not considered, as evidenced in Table~\ref{accuracy}.
	This underscores our method's potential for maintaining high accuracy after compression, consistent with the experimental findings presented in Table~\ref{accuracy_compression}.
	It is important to note that the anchor Bits Per Pixel (BPP) is set at 1, since transmitting a single spike incurs a cost of 1 bit, with the anchor accuracy detailed in Table~\ref{accuracy}.
	We observe a marginal decrease in accuracy (less than 20\% in most instances), while achieving a significant reduction in transmission bits, exceeding 95\% in savings.
	Furthermore, we employ the t-Distributed Stochastic Neighbor Embedding (t-SNE) technique~\cite{van2008visualizing} to reduce the dimensionality of predicted labels and visualize the outcomes, as depicted in Fig.~\ref{tsne}.
	Despite the compression, predicted labels for samples within the same category remain tightly clustered in the high-dimensional space, thereby ensuring high fidelity in classification accuracy.
	
	\begin{table}[]
		\small
		\renewcommand\arraystretch{1.3}
		\begin{tabular}{l|c|c|c|c}
			\bottomrule
			\multicolumn{1}{l|}{\multirow{2}{*}{\textbf{Type}}} & \multirow{2}{*}{\textbf{Method}} & \multicolumn{3}{c}{\textbf{Accuracy(\%)}}                                          \\
			\multicolumn{1}{l|}{}                      &                         & \multicolumn{1}{c|}{S-MNIST} & \multicolumn{1}{c|}{S-CIFAR} & S-CALTECH \\ \hline\hline
			\multirow{3}{*}{Event}                     & EtoF\cite{ahmad2022event}                    & 97.3                         & 58.5                         & 74.3         \\
			& BEI\cite{cohen2018spatial}                     & 89.6                         & 36.7                         & 51.8         \\
			& SBNE\cite{zhang2022discrete}                    & 96.2                         & 56.4                         & 73.8         \\ \hline
			\multirow{2}{*}{Spike}                     & SpiReco\cite{zhao2023spireco}                 & 99.1                         & 66.7                         & 73.1         \\ 
			& \textbf{Ours}                    & \textbf{99.6}                & \textbf{69.2}                & \textbf{77.4} \\ \bottomrule
		\end{tabular}
		\vspace{5pt}
		\caption{
			Comparison of classification accuracy between SOTA approaches for event and spike data, and our method.
			Accuracy of other approaches is obtained from~\cite{zhao2023spireco}.
			Results show that our method far exceeds other approaches, providing support for fidelity after compression.
		}
		\label{accuracy}
	\end{table}

	\begin{table}[]
		\centering
		\small
		\renewcommand\arraystretch{1.3}
		\begin{tabular}{c|l|c|c|c|c}
			\bottomrule
			Dataset                  & BPP              & 0.011     & 0.016     & 0.025     & 0.038     \\ \hline\hline
			\multirow{2}{*}{\textbf{S-MNIST}}   & BS(\%)     & 99.00 & 98.56 & 97.79 & 96.65 \\ \cline{2-6} 
			& AD(\%) & 13.2  & 12.0 & 11.6  & 10.8 \\ \hline
			\multirow{2}{*}{\textbf{S-CIFAR}}   & BS(\%)     & 98.76 & 98.18 & 97.14 & 95.66 \\ \cline{2-6} 
			& AD(\%) & 16.3      & 13.9      & 13.5      & 13.1      \\ \hline
			\multirow{2}{*}{\textbf{S-CALTECH}} & BS(\%)     & 98.91 & 98.41 & 97.53 & 96.25 \\ \cline{2-6} 
			& AD(\%) & 9.3      &  7.9     &  7.4     & 6.7     \\ \bottomrule
		\end{tabular}
		\vspace{5pt}
		\caption{
			Benefits of introducing compression to classification tasks.
			Comparisons are accomplished in bit-saving (BS) and accuracy-decreasing (AD) performance between our method on MNIST, CIFAR and CALTECH dataset.
			Results show that classification accuracy decreases slightly, while bit saving is significant, exceeding 95\%.
		}
		\label{accuracy_compression}
	\end{table}
	
	We present the classification report for the \textit{S-MNIST} dataset, as showcased in the Table~\ref{classification_report}.
	For simple characters such as 0, 5, and 7, our method demonstrates exceptional classification accuracy.
	However, it encounters challenges with more complex characters like 6, 8, and 9.
	In these cases, the model exhibits a tendency to erroneously categorize them as negative instances, leading to inaccuracies in the classification.
	This pattern suggests the need for further refinement in the model's ability to differentiate between characters with intricate features, underscoring an area for potential improvement in our approach.
		
	\begin{table}[]
		\centering
		\small
		\renewcommand\arraystretch{1.3}
		\begin{tabular}{l|c|c|c}
			\bottomrule
			\textbf{Category} & \textbf{Precision} & \textbf{Recall} & \textbf{F1-Score} \\ \hline \hline
			\textbf{0}            & 0.9883    & 0.9817 & 0.9849    \\ \hline
			\textbf{1}            & 0.9826    & 0.9883 & 0.9855    \\ \hline
			\textbf{2}            & 0.9818    & 0.9892 & 0.9855    \\ \hline
			\textbf{3}            & 0.9777    & 0.9850 & 0.9813    \\ \hline
			\textbf{4}            & 0.9617    & 0.9833 & 0.9724    \\ \hline
			\textbf{5}            & 0.9957    & 0.9642 & 0.9797    \\ \hline
			\textbf{6}            & 0.9693    & 0.9750 & 0.9722    \\ \hline
			\textbf{7}            & 0.9974    & 0.9592 & 0.9779    \\ \hline
			\textbf{8}            & 0.9529    & 0.9608 & 0.9568    \\ \hline
			\textbf{9}            & 0.9313    & 0.9492 & 0.9402    \\ \hline \hline
			\textbf{Accuracy}     & \multicolumn{3}{c}{0.9736}   \\ \hline
			\textbf{Macro Avg}    & 0.9739    & 0.9736 & 0.9736   \\ \hline
			\textbf{Weighted Avg} & 0.9739    & 0.9736 & 0.9736   \\ \bottomrule
		\end{tabular}
		\vspace{5pt}
		\caption{
			The classification report for \textit{S-MNIST} dataset.
			Results show that our method demonstrates exceptional classification accuracy for simple characters, while exhibiting a tendency to erroneously categorize for complex ones.
		}
		\label{classification_report}
	\end{table}
	
	\subsection{Ablation Study}
	We design experiments to illustrate the effectiveness of three technical contributions by visualizing results and features, including dual-pathway structure, FMVR and AFR.
	
\begin{figure}
		\centering
		\includegraphics[width=\linewidth]{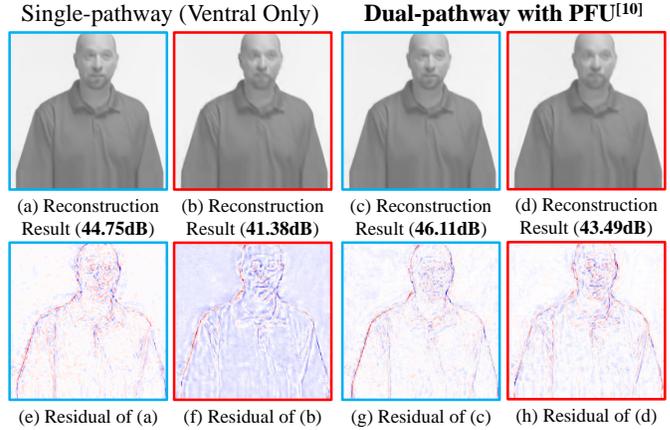}
		\caption{
			Visual comparison between reconstructed scenes via single- and dual-pathway structure.
			\textcolor{blue}{Blue} borders for low-speed scenarios and \textcolor{red}{red} borders for high-speed ones.
			Results show that dual-pathway structure and PFU exhibit greater gain for high-speed scenarios (2.11dB) compared to low-speed ones (1.36dB).
			Zoom in for better visibility.
		}
		\label{pfu_v}
	\end{figure}

	\textbf{Dual-pathway supervised structure.}
	The conventional single-pathway structure excels in capturing structural information for low-speed scenarios, enabling high quality scene reconstruction.
	However, it struggles to handle blur artifact caused by high-speed movement, which is shown in Fig.~\ref{pfu_v}(a) and (b).
	The proposed dual-pathway supervised structure adaptively captures motion characteristic, eliminating blur and discontinuity effect, which is shown in Fig.~\ref{pfu_v}(c) and (d).
	For scenarios with high-speed movement, the dorsal pathway extracts motion information such as direction and velocity, supervising the refinement of structural content through PFU.
	Results indicate that the method exhibits a greater gain in reconstructing high-speed scenes (2.11dB) compared to low-speed ones (1.36dB).
	By comparing Fig.~\ref{pfu_v}(f) and (h), it is learned that our proposed dual-pathway supervised structure can effectively reduce the deviation of overall light intensity within frame, enhancing the quality of reconstructed scene.
	As for Fig.~\ref{pfu_v}(e) and (g), it can be observed that blur located in edge areas is also significantly eliminated.
	Compared with ventral pathway only, the dual-pathway architecture improves PSNR results by \textbf{1.62dB} in average, demonstrating its effectiveness in handling spatial-temporal coupling information.
	
\begin{figure}
		\centering
		\includegraphics[width=.9\linewidth]{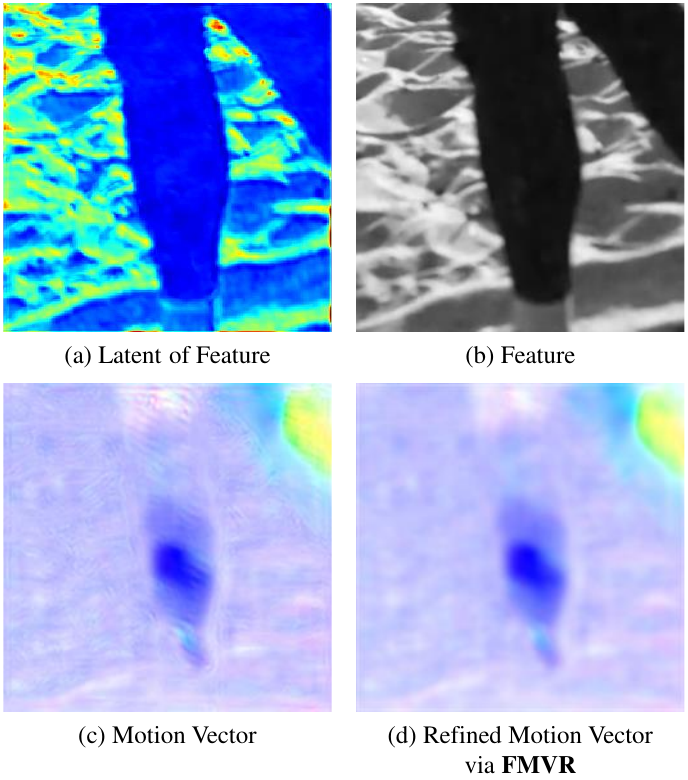}
		\caption{
			Visualization for (\textbf{a}) latent of feature, (\textbf{b}) feature, (\textbf{c}) motion vector and (\textbf{d}) refined motion vector.
			Results show FMVR eliminates molecular pattern which is discrete with feature content, supporting for precise alignment.
		}
		\label{refinement}
	\end{figure}

	\textbf{FMVR.}
	The warping operation in codec is performed in latent domain, making motion vectors similar with latents rather than features.
	It is observed that intensity is not uniformly distributed in regions containing rich textures, such as background area of Fig.~\ref{refinement}(a).
	This results in Moire artifact within the content, as illustrated in Fig.~\ref{refinement}(c).
	To ensure the consistency between motion vectors and features for more precise warping, we refine the motion vectors based on the content of features.
	The result shown in Fig.~\ref{refinement}(d) indicates that refined motion vector has eliminated texture and is similar with feature content.
	Through our proposed FMVR, the motion dynamic of distinct regions is enhanced such that Moire artifact is reduced.
	Regarding the reconstructed scene, the contextual content can be effectively aligned, which provides support for eliminating the ambiguity for intelligent tasks.
	The proposed FMVR module increases \textbf{1.22}, \textbf{0.03} and \textbf{0.01} for MUSIQ, DBCNN and CNNIQA respectively, indicating strong constraints on optical flow and feature content.
	
\begin{figure}
		\centering
		\includegraphics[width=.9\linewidth]{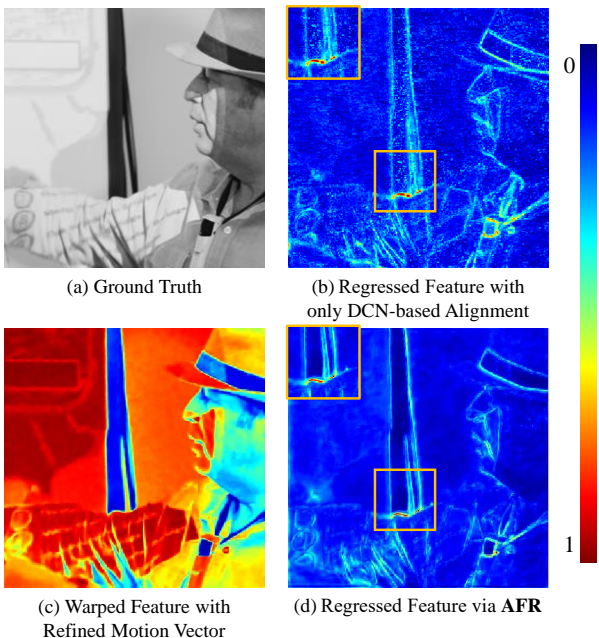}
		\caption{
			Visualization for (\textbf{a}) ground truth, (\textbf{b}) regressed feature with only PCD alignment, (\textbf{c}) warped feature with refined motion vector and (\textbf{d}) regressed feature with AFR.
			Results show that pixel-level inaccuracies are reduced and aligned feature exhibits smoother with assistance of warp-based alignment, leading to more accurate reconstructed scene.
		}
		\label{fau_v}
	\end{figure}

	\textbf{AFR.}
	Due to the limitation of receptive field, single DCN can only handle sub-pixel level offsets.
	Large-scale movement can be addressed by employing upsampling, as well as stacking hierarchical levels of DCN.
	However, low-resolution offsets may introduce pixel-level inaccuracies when upsampled to high resolution.
	This exceeds the processing capacity of DCN, resulting in pixel-level discontinuities after alignment which is shown in Fig.~\ref{fau_v}(b).
	We propose AFR to address this issue, using motion vector to represent large-scale movement.
	Warped with refined motion vector, scene can be essentially rebuilt, as illustrated in Fig.~\ref{fau_v}(c), which means motions with large-scale are well considered.
	This allows DCN to focus on capturing small-scale movement by capitalizing advantages in sub-pixel perception and generate precise offsets.
	After fusing results from warping and deformation, the regressed feature exhibits smoother transitions and leads to more accurate scene reconstruction, showing in Fig.~\ref{fau_v}(d).
	For non-reference assessments, metrics of method with AFR decreases \textbf{1.41}, \textbf{23.18} and \textbf{0.62} for NIQE, ILNIQE and PI respectively compared those of baseline, demonstrating its effectiveness in handling spatial-temporal coupling information.
	
	\begin{table*}[]
		\renewcommand\arraystretch{1.5}
		\setlength{\belowcaptionskip}{10pt}
		\centering
		\small
		\begin{tabular}{l|l|c|c|c}
			\bottomrule
			\textbf{Technical Contribution}                & \textbf{Approaches for Comparison} & \textbf{Flops} & \textbf{Parameters} & \textbf{Inference Time} \\ \hline \hline
			\multirow{3}{*}{\textbf{\begin{tabular}[c]{@{}l@{}}Dual-pathway and PFU\\ (End-Side)\end{tabular}}} & Extraction Module with Ventral Pathway Only      & 73.09G         & 1.11M & 31.44ms             \\ \cline{2-5} 
			& Extraction Module with Dual-pathway and PFU$^{[10]}$              & 74.79G         & 1.24M & 35.01ms             \\ \cline{2-5} 
			& Increasement                       & \textbf{2.33\%}         & \textbf{11.7\%} & \textbf{11.35\%}            \\ \hline
			\multirow{3}{*}{\textbf{\begin{tabular}[c]{@{}l@{}}FMVR\\ (End- \& Cloud-Side)\end{tabular}}}                 & Codec without FMVR                     & 585.55G        & 21.96M & 76.39ms            \\ \cline{2-5} 
			& Codec with FMVR                      & 585.85G        & 21.97M & 77.58ms            \\ \cline{2-5} 
			& Increasement                       & \textbf{0.05\%}         & \textbf{0.04\%} & \textbf{1.56\%}            \\ \hline
			\multirow{3}{*}{\textbf{\begin{tabular}[c]{@{}l@{}}AFR\\ (Cloud-Side)\end{tabular}}}                  & Analysis Module with DCN-based Alignment Only           & 288.30G         & 2.65M & 44.74ms             \\ \cline{2-5} 
			& Analysis Module with AFR                                & 464.77G         & 3.32M & 48.16ms             \\ \cline{2-5} 
			& Increasement                       & \textbf{61.21\%}        & \textbf{25.28\%} & \textbf{7.64\%}          \\ \hline
			
		\end{tabular}
		\vspace{5pt}
		\caption{
			Computational consumption comparison of technical contributions we proposed, including the Dual-pathway and PFU, FMVR and AFR.
			Results our technical contributions does not significantly increase the complexity and parameters at the end-side (\textbf{2.33\%} and \textbf{11.7\%}), while enhancing feature extraction performance and spike compression efficiency.
		}
		\label{complexity2}
	\end{table*}
	
	\begin{table*}[]
		\renewcommand\arraystretch{1.3}
		\setlength{\tabcolsep}{12pt}
		\centering
		\small
		\begin{tabular}{l|c|c|c|c|c|c}
			\bottomrule
			& \multicolumn{3}{c|}{SpikeCodec~\cite{feng2023spikecodec}}                                                                                          & \multicolumn{3}{c}{\textbf{Ours}}                                                                                 \\\cline{2-7}
			& \begin{tabular}[c]{@{}c@{}}Complexity\\ (GFlops)\end{tabular} & \begin{tabular}[c]{@{}c@{}}Parameter\\ (M)\end{tabular} & \begin{tabular}[c]{@{}c@{}}Inference Time\\ (ms)\end{tabular}& \begin{tabular}[c]{@{}c@{}}Complexity\\ (GFlops)\end{tabular} & \begin{tabular}[c]{@{}c@{}}Parameter\\ (M)\end{tabular} & \begin{tabular}[c]{@{}c@{}}Inference Time\\ (ms)\end{tabular} \\ \hline\hline
			\textbf{I-enc} & 679.40                                                        & 6.11                        & 155.57                           & \textbf{79.76}                                                         & \textbf{6.03}                &   \textbf{89.60}                                 \\
			\textbf{P-enc} & -                                                             & -              &-                                         & \textbf{96.13}                                                         & \textbf{3.90}                              &  \textbf{110.32}                    \\ \hline
			\textbf{I-dec} & 18.28                                                         & 42.88          & 60.59                                        & 457.96                                                        &  27.90                                     &  89.86            \\
			\textbf{P-dec} & -                                                             & -              &-                                         & 489.42                                                        & 18.06  & 97.67            \\ \bottomrule                                   
		\end{tabular}
		\vspace{5pt}
		\caption{
			Comparison of computational consumption between SpikeCodec~\cite{feng2023spikecodec} and our method.
			Results indicate that overhead for encoder is reduced significantly (88.26\%), providing support for spike encoding on low-power and embedded devices.
		}
		\label{complexity1}
	\end{table*}
	
	\subsection{Complexity Analysis}
	In order to thoroughly assess the impact of the three technical innovations we have proposed, we conduct a detailed comparison of the increases in complexity and parameters, which is systematically presented in  Table~\ref{complexity2}\footnote{
		The inference time is evaluated on Intel(R) i7-7700 @ 3.60GHz CPU and Geforce RTX 3080 GPU with input resolution of $256\times256$.
	}. 
	The results demonstrate that the dual-pathway architecture with PFU does not significantly increase the complexity and parameters (2.33\% and 11.7\%), while enhancing feature extraction performance and spike compression efficiency.
	On the other hand, while there is a notable increase in both complexity and parameters for AFR (61.21\% and 25.28\%), the inference time can be effectively reduced through parallel optimization techniques (only 7.64\% increasement), leveraging concurrent processing capabilities and allowing for more efficient execution of computational tasks.
	
	Since our framework is designed for the end-cloud collaborative environment, we further compare complexity and parameters of encoder (end-side) and decoder (cloud-side) between our method and SpikeCodec, which is the SOTA approach for reconstruction based spike compression, as shown in Table~\ref{complexity1}.
	\footnote{The inference time is evaluated on NVIDIA Geforce RTX 3080 and 4090 GPU for encoding and decoding respectively, simulating the computational difference between end-side and cloud-side.}
	Our scheme dramatically reduces computational consumption required for encoder, especially for complexity (88.26\%), which provides support for spike encoding on low-power and embedded devices~\cite{berthou2020accurate}\cite{tewolde2010current}.
	Despite the high overhead of decoder, the hardware resources in the cloud-side are fully affordable in end-cloud collaboration architecture, enabling real-time decoding and analysis.
	This clearly illustrates the superiority of our approach within the end-cloud collaborative framework, striking an effective balance between computational efficiency and performance enhancement.
	Our method significantly contributes to the popularization of spike cameras and spike-based applications.
	
	It's worth mentioning that, the analysis model deployed at the cloud-side can be adjusted for different tasks, and the model size can also be adapted according to granularity and precision requirements, demonstrating the generality, flexibility, and robustness of our approach.
	Future research endeavors should concentrate on two key aspects. Firstly, enhancing the parallel processing capabilities of the feature extraction module is crucial. This improvement aims to reduce inference time, thereby enabling real-time feature extraction and compression. Secondly, it is imperative to simplify the complexity of the analytical network, decreasing both the hardware requirements and the associated costs, making the technology more efficient. These focal points will collectively contribute to the development of more agile and cost-effective systems, suitable for a wider range of applications.
	

	\section{Conclusion}
	In this paper, we for the first time conceptualize the SCI, which jointly optimizes the low-level spike compression and high-level spike analysis.
	We propose a novel spike compression framework encapsulated with dual-pathway structure, advanced motion expression module and feature regression module.
	Our method realizes high efficiency spike compact representation and high performance spike reconstruction and classification with low computational resource consumption.
	Extensive experimental results show that our scheme obtains SOTA performance in both compression and analysis tasks.
	This research enables a novel sub-field for spike visual intelligence, facilitating future study.

	\bibliographystyle{unsrt}
	\bibliography{tip}
	
\end{document}